# PeSOA: Penguins Search Optimisation Algorithm for Global Optimisation Problems

Youcef GHERAIBIA[1], Abdelouahab MOUSSAOUI[2], Peng-Yeng YIN[3], Yiannis PAPADOPOULOS[1], , and Smaine Maazouzi[4]

[1]School of Engineering and Computer Science, University of Hull, U.K

[2]Department of Computer science, University of Setif 1, 19000, Algeria

[3]Department of Information Management, National Chi Nan University, Taiwan

[4]Department of Computer Science, Univsersité 20 Août 1955, Algeria

**Abstract:** *This paper develops PeSOA, a new metaheuristic algorithm which is inspired by the foraging behaviours of penguins. A population of penguins located in the solution space of the given search and optimisation problem is divided into groups and tasked with finding optimal solutions. The penguins of a group perform simultaneous dives and work as a team to collaboratively feed on fish the energy content of which corresponds to the fitness of candidate solutions. Fish stocks have higher fitness and concentration near areas of solution optima and thus drive the search. Penguins can migrate to other places if their original habitat lacks food. We identify two forms of penguin communication both intra-group and inter-group which are useful in designing intensification and diversification strategies. An efficient intensification strategy allows fast convergence to a local optimum, whereas an effective diversification strategy avoids cyclic behaviour around local optima and explores more effectively the space of potential solutions. The proposed PeSOA algorithm has been validated on a well-known set of benchmark functions. Comparative performances with six other nature-inspired metaheuristics show that the PeSOA performs favourably in these tests. A run-time analysis shows that the performance obtained by the PeSOA is very stable at any time of the evolution horizon, making the PeSOA a viable approach for real world applications.*

## 1. Introduction

Nature-inspired metaheuristic approaches have been applied to solve NP-hard problems such as parameter estimation [23], vehicle routing problems [24], the traveling salesmen problem [27], Dynamic Deployment of Wireless Sensor Networks [31], and to bioinformatics [22]. Notable metaheuristics include genetic algorithms (GA) [4], differential evolution (DE) [25], particle swarm optimisation (PSO) [14], ant colony optimisation (ACO) [3], artificial bee colony (ABC) [15], firefly algorithms (FA) [28], cuckoo search (CS) [29], bat algorithms (BA) [30], simulated annealing (SA) [1], Tabu search (TS) [10], scatter search (SS) [16], and The greedy randomised adaptive search procedure (also known as GRASP) [5]. These metaheuristics can be classified according to different search characteristics such as Trajectory-based methods, Population-based methods and Memory usage. Two major search strategies have been largely taken into account in pursuing optimisation. The intensification strategy aims to exploit previously-found promising regions in order to detect local optima. The diversification strategy strives to explore uncharted regions to identify new trajectories that might lead to the global optimum. The two strategies work in cooperation to iteratively improve the best solution on hand.

Nature inspired metaheuristics have demonstrated success in a plethora of problems and applications. However, there is always space to explore new techniques that draw inspiration from nature in the hope that more effective and efficient heuristics can be devised. The Penguin search Optimisation Algorithm (PeSOA) is such a novel population- and memory-based metaheuristic approach which was first proposed in [8]. PeSOA is inspired by the penguin's hunting behaviour and it generally works as follows. The population of penguins locates initial positions, each penguin then dives and swims under the water for hunting fish while consuming its oxygen reserve. Different forms of the communication between penguins are occasionally performed and the quantities of eaten fish increase. The process is repeated until the specified amount of fish is obtained or the maximum number of iterations is reached. [8] have shown that the PeSOA outruns genetic algorithms and particle swarm optimisation in obtaining better values for benchmark optimisation functions. This paper enhances PeSOA with improved intensification and diversification strategies. The main differences between the present work and the original PeSOA are as follows.

In the present work, the penguins are dispatched into several groups where each group is allotted to a separate region in the food space. During the foraging phase, the penguins of each group attempt to hunt a maximal number of fish around the allotting region. The communication for sharing food information happens inter- and intra-group, allowing to improve the best solution on hand. We have codified PeSOA in java and tested the algorithm with a well-known set of benchmark global optimisation functions. The results reveal that the proposed approach outperforms a prominent set of the-state-of-the-art bio-inspired metaheuristics, including GA, DE, PSO, ABC, and BA.

The rest of this paper is organised as follows. Section 2 describes the hunting behaviour of penguins. Section 3 articulates the proposed PeSOA algorithm. Section 4 presents the experimental results including the comparative performances. Finally, we conclude the paper by some remarks and future perspectives in Section 5.

## 2. Metaphor: Hunting Behaviour of Penguins

Penguins are sea birds and they are unable to fly because of their adaptation to aquatic life [9], [12], [27], [21]. Their wings are ideal for swimming and can be considered as fins. Penguins remain under the water for up to twenty minutes so they can go deeper. Penguins can dive more than 520m to scan the water for food. Although it is more efficient and less tiring to swim under the water than slithering on the ice, they must regularly surface every couple of minutes for air. They are able to breathe while swimming rapidly (7 to 10 km/h) [26] by slowing down the heart rate and keeping their eyes open for scanning food. The retina allows penguins to distinguish shapes and colours. Penguins feed on krill, small fish, squid, and crustaceans. It takes up more energy for them to dive deeper and longer, so they have to consume more food this way.

The optimisation of foraging behaviour was modelled in the works of [17], [18]. These two studies hypothesised that dietary behaviour may be explained by the economic reasoning: it comes to a profitable food search activity when the gain of energy is greater than the expenditure required to obtain this gain. Penguins, behaving along the line of foraging predators, must extract information about the time and cost to get food and the energy content of prey in order to choose the course for making their next dive. The air-breathing behaviour of penguins was noticed by [11]. The land is a home base for penguins who are forced to surface for air after each foraging trip. A trip implies immersion in apnea. The duration of a trip is limited by the oxygen reserves of penguins, and the speed at which they use it [13], [27].

For saving the energy and the oxygen reserves, penguins must feed as a team and synchronise their dives to optimise the foraging. Penguins communicate with each other with vocalisations. These vocalisations are unique to each penguin (like fingerprints to humans). Therefore, they allow the unique identification and recognition for penguins between each other [19].

## 3. The PeSOA Algorithm

To summarise the observations from penguins' foraging behaviour, the following rules are presented.

**Rule 1:** A penguin population comprises of several groups. Each group contains a number of penguins that varies depending on food availability in the corresponding foraging region.

**Rule 2:** Each group of penguins starts foraging in a specific depth under the water according to the information about the energy gain and the cost to obtain it.

**Rule 3:** They feed as a team and follow their local guide which has fed on most food in the last dive. Penguins scan the water for food until their oxygen reserves are depleted.

**Rule 4:** After a number of dives, penguins return on surface to share with its local affiliates, via intra-group communication, the locations and abundance of food sources.

**Rule 5:** If the food support is less for the penguins of a given group to live on, part of the group (or the whole group) migrates to another place via inter-group communication.

In Table 1, we relate these rules to principles of optimisation heuristics. The sea corresponds to the solution space and the goal of the penguin searching is to locate the best position under the water showing the most abundant shoals of fish. The position of each individual penguin is thus a candidate solution to the optimisation problem. The energy of the penguin obtained by catching prey in terms of the quantity of fish around a position is analogous to the fitness of the solution. The oxygen reserve of a penguin reflects its health condition that serves as an acceleration coefficient in an instance of swimming. Finally, the two forms of communication represent the metaheuristic search strategies to increase the likelihood for targeting the global optimum.

Table 1. Metaphors of penguin hunting behaviours for optimisation heuristic principles.

| Penguin hunting behaviours | Optimisation heuristic principles |
|---|---|
| The sea | Solution space |
| Most abundant shoals of fish | Global optimum |
| Penguin position | A candidate solution |
| Energy content of prey | Fitness of a solution |
| Oxygen reserve | Acceleration coefficient |
| Penguin swimming | Solution update |
| Intra-group communication | Intensification search |
| Inter-group communication | Diversification search |

The behavioural ecology of penguin foraging is in many ways similar to modern metaheuristics. This nature intelligence has inspired us for developing the penguin search optimisation algorithm (PeSOA). The general ideas of the PeSOA work as follows. The penguins are divided into groups (not necessarily with the same cardinality) and each group starts foraging with a specific region. The status of each penguin is represented by its position and oxygen reserve. After a number of dives, the penguin returns to surface and share with its group affiliates the position and quantity of the food found.

The local best of each group continuously improves as more members report the food sources. After an entire cycle of the intra-group communication of all the penguin groups, the penguins might migrate to other group's habitat according to the probability of nurture existence of each group in terms of the quantity of food found by all its members. The collaboration of team foraging repeats until a maximal number of cycles have been performed. With the notations defined in Table 2, the searching heuristics performed by the PeSOA are articulated as follows.

Table 2. Notation descriptions.

| Notations | Descriptions |
|---|---|
| N | Number of total penguins |
| K | Number of groups |
| f | Objective function of the problem |
| $O_j^i$ | The oxygen reserve of the $j^{th}$ penguin of the $i^{th}$ group |
| $x_j^i(t)$ | The position of penguin j allocated to the $i^{th}$ group at $t^{th}$ instance |
| $x_{LocalBest}^i$ | The best solution found by the $i^{th}$ group |
| $QEF^i(t)$ | Quantity of eaten fish of the $i^{th}$ group at the $t^{th}$ instance |
| $P_i(t)$ | Probability of existence of fish of $g^{th}$ group at $t^{th}$ instance |
| rand() | A random number drawing from (0, 1) |

### 3.1 Swimming course update

Let G = {G$_1$, G$_2$…G$_K$} be the set of K disjoint groups of penguins randomly distributed in the whole solution space Ω. Each group **G$_i$** contains **d$_i$** penguins where each penguin j in **G$_i$** is placed at a solution at time instance t, the penguin j swims to a new position at time t+1 in Ω by the following expression.

$$x_j^i(t+1) = x_j^i(t) + O_j^i(t) \times rand() \times \left(x_{LocalBest}^i - x_j^i(t)\right) \quad (1)$$

Equation (1) can be realised by penguin swimming behaviour. Penguins primarily rely on their vision while hunting. Penguins follow their local leader who has found most food in the last dive, and they explore the along the path guided by the local leader. The penguin swimming is accelerated by the oxygen reserve which reflects its health condition determined by previous dives. In terms of optimisation terminology, the trial solution is updated by moving towards the local best solution with a random turbulence. The moving distance depends on the acceleration coefficient which is a variable adapted by previous gains along the pursued trajectory. If the solution keeps ameliorated, indicating a promising direction of the trajectory, the value of the acceleration coefficient increases and promotes a great moving distance.

### 3.2 Oxygen reserve update

After each dive, the oxygen reserve of the penguin is updated as follows.

$$O_j^i(t+1) = O_j^i(t) + \left(f(x_j^i(t+1)) - f(x_j^i(t))\right) \times \left\|x_j^i(t+1) - x_j^i(t)\right\| \quad (2)$$

Where f is the objective function of the underlying problem. The oxygen reserve depends on both the gain of the food source and the swimming duration the penguin endures. If the energy gain is positive, the longer the penguin stays under the water, the more quantities of food it catches and thus becomes healthier. Otherwise, the longer the swimming duration, the more oxygen the penguin consumes. Hence, the oxygen reserve is updated according to the amelioration of the objective function. The oxygen reserve increases if the new solution is better than the previous one, and the oxygen reserve decreases in the other case. The penguin performs repetitive dives until the oxygen is depleted, then the penguin will migrate to another group due to the undersupply of food in this area.

### 3.3 Intra-group communication

Penguins feed on food as a team and they manage well intra-group communication. Penguins follow the local guide who made the most successful trial in the last dive (see Equation (1)). For every instance of dive, the penguin may find a better food source and becomes the new local guide. The team foraging is an autocatalytic process which assures the continuous amelioration of trial solutions.

### 3.4 Food abundance update

The food abundance degree associated to a group indicates the energy content of prey captured by all the members in that group (see Equation (3)). In the light of penguin foraging, the food abundance degree can be estimated by the Quantity of Eaten Fish (QEF), which is calculated by the following expression.

$$QEF^i(t+1) = QEF^i(t) + \sum_{j=1}^{d_i}\left(O_j^i(t+1) - O_j^i(t)\right) \quad (3)$$

The QEF of a group represents the attractiveness the penguin members would stick to that group. A great QEF value means the region affords enough food for the whole group and even solicits penguins migrating from other groups.

### 3.5 Group membership update

The penguin may migrate to join another group due to the food undersupply in the group it originally belongs. The penguin updates its group membership by reference to a function relating to the food abundance degree of various groups. The penguin joins a group with a probability proportional to the corresponding group's QEF, increasing the success likelihood of food foraging in the next dive. On the other hand, the region explored by a group is abandoned if all the members of that group have migrated to other groups. The membership function value of joining the group i is a probability given as follows.

$$P_i(t+1) = \frac{QEF^i(t)}{\sum_{j=1}^{K} QEF^j(t)} \quad (4)$$

Hence, the inter-group communication facilitates a form of proportional biased diversification search capability, the promising region containing more abundant food would be intensively contemplated by augmenting the number of group members. In terminology of evolutionary computation, the penguin's inter-group communication resembles to the survival of the fittest genes that provides the building blocks for constructing quality solutions.

### 3.6 Pseudo code of PeSOA

The pseudo code of the penguin search optimisation algorithm (PeSOA) is shown in Algorithm 1. As will be noted, two search strategies (Algorithm 2 and Algorithm 3) originally proposed in scatter search [15] are used to enhance the performance of PeSOA. In summary, the PeSOA starts with K diversified groups of penguins. Each penguin searches for food separately in its assigning group with the guidance of the local best solution. After each cycle of dives, penguins of the same group share with each other the information about the position and quantity of the food. When the oxygen reserve is depleted, the penguin returns to surface and share the group information with members from the other groups. Then, the penguin is redistributed according to the updated group membership function. The search process is repeated until the stopping criterion is reached. The diversification generation strategy (see Algorithm 2) is used to generate K diversified groups in the initial penguin population. PeSOA starts with a population distributed in K groups, and each group is placed in a separate region with a minimum distance to any other. The purpose is to start the search with a set of diversified initial solutions which have contrasting features benefiting in future solution improvement. The solution improvement strategy (see Algorithm 3) is used to lead the penguin swimming to a local optimum after performing a complete cycle of dives. This is a common practice in modern metaheuristics, such as GRASP or hybrid GAs, where a local search component is embedded in the evolutionary cycle in order to utilise the key building blocks contained in local optima. The penguin swimming is guided by the local best solution for the group and accelerated by the oxygen reserve. The oxygen reserve indicates the health condition of the penguin. The penguins with a high reserve of oxygen have a good energy-ameliorating path, which then promotes the penguins to last longer in the water and swim a greater distance towards the same direction. If the penguin ameliorates its objective function value in this dive, the penguin solution is updated and the local best solution is also checked for possible update. If the penguin fails to find a better food source in this dive, its position is not changed, however, the oxygen reserve is still being updated due to the oxygen consumption in performing this dive.

**Algorithm 1: Algorithm of PeSOA**
1: **Generate** K regions in the solution space with **Algorithm 1**;
2: **Generate** penguins $x_j^i$ (j = 1, 2, ..., N/K) for each group i within the designated region;
3: **while** stopping criterion is not reached **do**
4: **Initialise** the oxygen reserve for each penguin;
5: **For** each group i **do**
6: **For** each penguin j in this group **do**
7: **Improve** the penguin position $x_j^i$ with **Algorithm 2**;
8: **End**
9: **Update** the food abundance degree for this group by **Eq. (3)**;
10: **End**
11: **Update** the global best solution;
12: **Update** membership function values for each group by **Eq. (4)**;
13: **Redistribute** penguins to groups according to the membership function;
14: **Abandon** the group if it has no members;
15: **end while**
16: **End**

**Algorithm 2: Diversification generation strategy**
1: **Input**: Solution space, K (number of groups); MaxDist (minimum inter-group distance).
2: **Output**: K region centers in the solution space
3: **choose** the center of the first group randomly, denoted by $C_0$
4: i ← 1;
5: **while** i < K **do**
6: **choose** a center $C_i$ randomly for the next group
7: j ← 0;
8: **while** j < i **do**
9: **if** Distance($C_i$, $C_j$) > MaxDist **then**
10: j ← j + 1
11: **else** go to step 6
12: **end if**
13: **end while**
14: i ← i+1;
15: **end while**
16: **End**

**Algorithm 3: Solution improvement strategy**
1: **Input**: $x_j^i$, $O_j^i$, $x_{LocalBest}^i$
2: **Output**: new $x_j^i$, $x_{LocalBest}^i$
3: **while** $O_j^i > 0$ **do**
4: **Take** a dive for $x_j^i$ according to **Eq. (1)**
5: **if** $x_j^i$ improves **then**
6: **Update** $x_j^i$
7: **Update** $x_{LocalBest}^i$ if $x_j^i$ beats $x_{LocalBest}^i$
8: **end if**
9: **Update** $O_j^i$ using **Eq. (2)** //no matter if $x_j^i$ has been updated or not)
10: **end while**
11: **End**

### 3.7 Computational Complexity of PeSOA

The Algorithm PeSOA is divided into two parts, the generation of initial population (step 1 and step 2) and the iterative evolutionary search (step 3 to step 15).

The computational complexity of the first part is $O(K^2+N)$. The computational complexity of the second part is $O(N \times t)$ where $t$ is the maximum number of evolutionary iterations. The overall complexity of the Algorithm PeSOA is thus $O(K^2 + N \times t)$ which is comparable to that for most nature-inspired metaheuristics.

## 4. Experimental Results

### 4.1 Parameter Settings

The application of PeSOA requires appropriate settings of the critical parameters such as the number of groups and the penguin population size. The parameter values are often chosen heuristically due to the fact that the determination of the optimal parameter values is itself can NP-hard problem. We propose to find the best PeSOA parameter values by maximising the ratio between the mean gain in objective function amelioration and the mean consumed CPU time. We test the PeSOA with five benchmark functions (Ackley, Sphere, Rastrigin, Rosenbrock, and Griewank functions) for a sufficient number of instances for each parameter. Fig. 1 shows the performance ratio of the PeSOA against the number of groups ranging from 2 to 50. We observe, for all test functions in general, that the best performance ratio is obtained when the penguins are initially distributed to about five groups. Similarly, the PeSOA is tested against the number of penguins initially assigned to each group ranging from 5 to 100 with an increment of five penguins. It is seen that the performance ratio of the PeSOA reaches the best value when the group size is between 40 to 50 penguins.

### 4.2 Test and Validation

In the literature a set of benchmark functions [20] has been intensively used to test and validate metaheuristic algorithms. These benchmark functions express diverse criteria to verify the characteristics of the optimisation algorithms such as robustness, sensitivity, and scalability. Table 3 describes the information for these benchmark functions, including the function name, number of decision variables (D), function expression, global optimum, and the variable bound. We compare the PeSOA with several the-state-of-the-art nature-inspired metaheuristics, including PSO, ABC, BA, GA, DE, and CS. The values shown are the means and the standard deviations over ten independent runs of each algorithm. It is seen that the PeSOA obtains the best mean objective value for twenty functions (F01-F06, F09, F10, F15, F16, F17, F18, and F20). For the rest of the benchmark functions, the PeSOA also obtains comparable objective values to those by other competing algorithms. The names of each function is (F 01,Hartman 1),(F 02,Hartman 2), (F 03,Kowalik),(F 04,Shekel 1),(F 05,Shekel 2),(F 06,Shekel 3),( F 07,Branin),(F 08,Ackley),(F 09, Griewank 10),( F 10 ,Griewank 20),(F 11, Griewank 30),( F 12, Quartic noise),( F 13, Rastrigin 10),( F 14, Rastrigin 20),( F 15, Rastrigin 20),( F 16, Rosenbrock 10),( F 17, Rosenbrock 20),( F 18, Rosenbrock 30),( F 19, Schwefel 2.26),( F 20,Sphere).

Table 4 reports the mean CPU running time consumed by these algorithms for obtaining the previously noted objective values. We see that the PeSOA consumes the least CPU running time for all the test functions. In summary, the PeSOA serves as the most effective and efficient algorithm in terms of the performance ratio between the mean gain in objective function amelioration and the mean consumed CPU time. As the nature-inspired metaheuristics are stochastic optimisation algorithms, each independent run of the same algorithm may manifest distinct run-time behaviours compared to other runs. It is thus very crucial to analyse the variation the best obtained function value as the number of used function evaluations increases.

## 5. Concluding Remarks and Future Works

In this paper we have developed a new meta-heuristic algorithm for global optimisation. The new approach is based on the collaborative foraging strategy applied by penguins. The oxygen reserve of the penguin, indicating its health condition, is used to control the swimming step size and the length of the duration the penguin stays under the water. The group local best solution is used to guide the penguin members to generate new solutions. The penguins will migrate to other groups if its original assigning group is unable to afford enough food. The proposed PeSOA algorithm is validated on a well-known set of benchmark functions broadly used in the literature, and a performance comparison is made with several nature-inspired metaheuristic algorithms such as PSO, ABC, DE, GA and BA. Simulation results showed that the PeSOA is more robust and efficient compared to other competing algorithms because its search strategy does not rely only on changing the next position of the best solution found, but also on penguin communication happening both within and between groups. The original PeSOA algorithm has been used to solve combinatorial problems such as automotive safety integrity levels allocation [7], Capitated vehicle routing problem [2] and optimal spaced seed finding [6]. The PeSOA algorithm can be extended in several ways, for example, the introduction of reproduction and migration may enhance the search capability. It is worth studying the multi-objective version of the PeSOA.

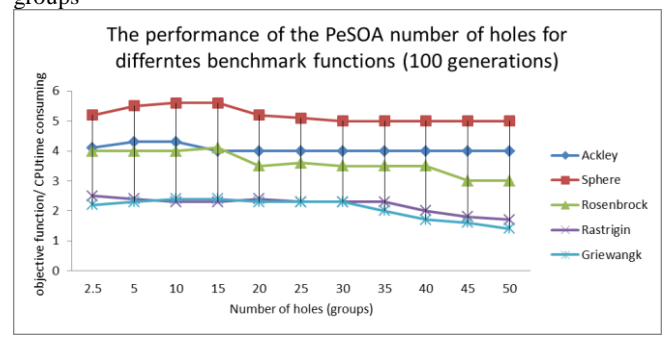

Fig. 1. The performance of the PeSOA against the number of groups

Table 3. The best function values obtained by competing algorithms. The values shown are the means and the standard deviations over ten independent runs of each algorithm.

| F | PSO | ABC | BA | GA | DE | PeSOA |
|---|---|---|---|---|---|---|
| F 01 | -3.6384 (±4.44 e-003) | -3.7197 (±5.0073 e-003) | -3.6294 (±1.0578 e-002) | -3.5591 (±3.7758 e-003) | -3.7004 (±3.2557 e-003) | -3.8597 (±1.1027 e-004) |
| F 02 | -3.2108 (±0.1057) | -3.2879 (±0.0546) | -3.2519 (±0.2349) | -3.1908 (±0.10250) | -3.2290 (±0.06957) | -3.3194 (±0.00108) |
| F 03 | -2.7306 e-4 (±0.0002 e-4) | -3.0001 e-4 (±0.0006 e-4) | -2.9001 e-4 (±0.0009 e-4) | -2.7480 e-4 (±0.0006 e-4) | -2.9979 e-4 (±0.00013 e-4) | -3.008 e-4 (±0.0003 e-4) |
| F 04 | -9.6811 (±9.4361 e-003) | -9.9073 (±2.7326 e-003) | -10.0120 (±4.1197 e-002) | -9.8107 (±4.6228 e-003) | -10.1009 (±4.2106 e-002) | -10.1508 (±1.0214 e-003) |
| F 05 | -9.8788 (±3.2643 e-004) | -10.2638 (±3.1271 e-004) | -10.1911 (±2.1975 e-005) | -9.8389 (±1.0674 e-004) | -10.2374 (±0.1164 e-003) | -10.3867 (±3.10247 e-003) |
| F 06 | -10.1915 (±9.3467 e-004) | -10.4700 (±8.0468 e-004) | -10.4108 (±3.6234 e-003) | -10.0527 (±4.9346 e-004) | -10.3893 (±1.0017 e-002) | -10.5104 (±3.1027 e-003) |
| F 07 | 0.3999 (±2.6227 e-004) | 0.3927 (±8.6794 e-004) | 0.3967 (±2.03647 e-004) | 0.4035 (±8.0637 e-003) | 0.3901 (±1.0037 e-002) | 0.38794 (±3.27680 e-002) |
| F 08 | 1.9734 e-005 (±2.8745 e-007) | 1.6012 e-005 (±0.0101 e-005) | 1.7845 e-005 (±2.0012 e-006) | 2.0324 e-005 (±1.0327 e-006) | 1.3024 e-005 (±0.0148 e-005) | 1.0001 e-005 (±1.0067 e-006) |
| F 09 | 3.9987 e-007 (±0.9534 e-008) | 3.8634 e-007 (±0.9254 e-008) | 3.8374 e-007 (±0.7934 e-008) | 3.9222 e-007 (±0.1423 e-008) | 3.8145 e-007 (±0.5214 e-008) | 3.6874 e-007 (±0.0329 e-008) |
| F 10 | 6.2547 e-007 (±0.0213 e-007) | 6.0012 e-007 (±0.2111 e-007) | 6.0669 e-007 (±0.1992 e-007) | 6.9117 e-007 (±0.1423 e-008) | 6.0001 e-007 (±0.5398 e-008) | 5.9998 e-007 (±0.0014 e-007) |
| F 11 | 1.1404e-006 (±0.0207 e-008) | 1.6349 e-006 (±0.0507 e-008) | 9.99267 e-007 (±0.0067 e-008) | 1.3684e-006 (±0.4954 e-008) | 1.6658 e-006 (±0.6279 e-008) | 9.7238 e-007 (±0.0103 e-008) |
| F 12 | 1.4755 (±0.98542 e-02) | 1.29215 (±0.7649 e-03) | 1.14291 (±0.3167 e-02) | 1.1458 (±0.6145 e-02) | 1.3765 (±0.8654 e-02) | 1.00859 (±0.00215 e-02) |
| F 13 | 3.6230 e-08 (±0.5214 e-09) | 3.4010 e-08 (±0.0987 e-09) | 3.6005 e-08 (±0.2301 e-10) | 3.8254 e-08 (±0.2124 e-08) | 3.9103 e-08 (±0.1038 e-07) | 3.0678 e-08 (±0.0038 e-09) |
| F 14 | 6.2017 e-08 (±0.1120 e-08) | 6.3980 e-08 (±0.2303 e-08) | 6.2097 e-08 (±0.4921 e-9) | 6.5325 e-08 (±0.4009 e-09) | 6.1410 e-08 (±0.4947 e-08) | 5.0101 e-08 (±0.1934 e-08) |
| F 15 | 6.3254 e-08 (±0.6291 e-10) | 6.0034 e-08 (±0.9004 e-10) | 6.2015 e-08 (±0.8074 e-10) | 6.8354 e-08 (±0.2094 e-10) | 6.9574 e-08 (±0.3005 e-10) | 5.9887 e-08 (±0.1027 e-09) |
| F 16 | 9.4352 e-10 (±4.6349 e-12) | 9.854 e-10 (±2.3276 e-11) | 9.0136 e-10 (±0.0362 e-10) | 9.3492 e-10 (±0.2934 e-10) | 8.9653 e-10 (±0.5731 e-12) | 6.16587 e-10 (±2.0374 e-9) |
| F 17 | 1.0321 e-9 (±0.2024 e-10) | 1.1124 e-9 (±0.0094 e-9) | 1.2000 e-9 (±0.0164 e-9) | 1.1162 e-9 (±0.6216 e-10) | 1.2110 e-9 (±0.3140 e-11) | 1.0064 e-9 (±0.0491 e-10) |
| F 18 | 1.3254 e-09 (±1.1637 e-12) | 2.0014 e-09 (±0.9175 e-10) | 1.0153 e-09 (±2.014 e-11) | 1.8632 e-09 (±1.1124 e-10) | 0.9843 e-09 (±1.0310 e-11) | 0.9012 e-09 (±0.0248 e-9) |
| F 19 | -9.8994 e+3 (±8.0349 e-02) | -9.8497 e+3 (±3.4328 e-02) | -9.8346 e+3 (±3.0028 e-02) | -9.9934 e+3 (±1.9919 e-03) | -9.6527 e+3 (±0.3015 e-02) | -9.6254 e+3 (±1.3756 e-02) |
| F 20 | 5.9648 e-006 (±7.0364 e-008) | 5.8555 e-006 (±0.6249 e-007) | 5.5364 e-006 (±9.3248 e-007) | 5.8774 e-005 (±2.0317 e-007) | 5.8749 e-006 (±0.1367 e-007) | 5.4237 e-006 (±2.1038 e-009) |

Table 4. The CPU running time consumed by competing algorithms. The values shown are the means and the standard deviations over ten independent runs of each algorithm.

| F | PSO | ABC | BA | GA | DE | PeSOA |
|---|---|---|---|---|---|---|
| F 01 | 1.0937(± 0.0060) | 0.9375(± 0.0092) | 1.0006 (± 0.0035) | 1.0937 (± 0.0060) | 0.9976 (± 0.0111) | 0.8134(± 0.0013) |
| F 02 | 1.2531(± 0.0108) | 1.1120(± 0.0063) | 1.2193 (± 0.0062) | 1.3012 (± 0.0094) | 1.1865 (± 0.0039) | 1.0014(± 0.0301) |
| F 03 | 2.5092(± 0.1002) | 2.1834(± 0.1108) | 2.3321 (± 0.0974) | 2.5248 (± 0.0324) | 2.2016 (± 0.0010) | 1.9937 (± 0.0827) |
| F 04 | 1.2843(± 0.0162) | 0.9734(± 0.0019) | 1.0999 (± 0.0085) | 1.3012 (± 0.0098) | 0.9874 (± 0.0022) | 0.9329 (± 0.0567) |
| F 05 | 1.2019(± 0.0083) | 1.1376 (± 0.0087) | 1.1248 (± 0.0082) | 1.1987 (± 0.0036) | 1.0875 (± 0.0019) | 0.8934 (± 0.0627) |
| F 06 | 1.2354(± 0.0079) | 1.1018(± 0.0052) | 1.1364 (± 0.0029) | 1.2402 (± 0.0018) | 1.0985 (± 0.0130) | 0.9978 (± 0.0136) |
| F 07 | 0.5362 (± 0.0031) | 0.4501(± 0.0082) | 0.4737 (± 0.0022) | 0.5408 (± 0.0013) | 0.4700 (± 0.0008) | 0.4237 (± 0.0238) |
| F 08 | 0.9514(± 0.0034) | 0.7924 (± 0.0068) | 0.8350 (± 0.0133) | 0.9724 (± 0.0021) | 0.8013 (± 0.0003) | 0.8054 (± 0.0346) |
| F 09 | 0.8625(± 0.0001) | 0.7962 (± 0.0068) | 0.8070 (± 0.0046) | 0.88137± 0.0019) | 0.7924 (± 0.0108) | 0.6854(± 0.0349) |
| F 10 | 0.8894(± 0.0002) | 0.8994 (± 0.0021) | 0.8545 (± 0.0033) | 0.9000(± 0.0009) | 0.8421 (± 0.0821) | 0.7998(± 0.0800) |
| F 11 | 0.9012(± 0.0014) | 0.8635 (± 0.0106) | 0.8832 (± 0.0002) | 0.9235(± 0.0150) | 0.8436 (± 0.0301) | 0.8001(± 0.0131) |
| F 12 | 1.3628(± 0.0324) | 1.1083 (± 0.0090) | 1.2651 (± 0.0131) | 1.4132(± 0.0134) | 1.2107 (± 0.0318) | 0.9241 (± 0.0043) |
| F 13 | 0.6192(± 0.0061) | 0.5000(± 0.0008) | 0.5579 (± 0.0049) | 0.6301 (± 0.0107) | 0.5132 (± 0.0087) | 0.4135(± 0.0010) |
| F 14 | 0.6882(± 0.0009) | 0.5865(± 0.0090) | 0.6237 (± 0.0101) | 0.6709 (± 0.0034) | 0.5939 (± 0.0010) | 0.4821(± 0.0087) |
| F 15 | 0.7014 (± 0.0100) | 0.6210(± 0.0032) | 0.5920 (± 0.0092) | 0.7011 (± 0.0091) | 0.6008 (± 0.0031) | 0.5010(± 0.0010) |
| F 16 | 0.7014(± 0.0083) | 0.4987(± 0.0005) | 0.5521 (± 0.0063) | 0.7301 (± 0.0030) | 0.5214 (± 0.0103) | 0.4937 (± 0.0013) |
| F 17 | 0.7541(± 0.0109) | 0.5214 (± 0.0010) | 0.5771 (± 0.0139) | 0.7634 (± 0.0009) | 0.5635 (± 0.0019) | 0.5118 (± 0.0108) |
| F 18 | 0.7924 (± 0.0012) | 0.5674 (± 0.0010) | 0.5991 (± 0.0192) | 0.7994 (± 0.0012) | 0.6014 (± 0.0019) | 0.5384 (± 0.0089) |
| F 19 | 1.3287(± 0.0529) | 1.2104(± 0.0107) | 1.2768 (± 0.0044) | 1.3762(± 0.0130) | 1.2001 (± 0.0009) | 1.0034 (± 0.0010) |
| F 20 | 0.6294(± 0.0017) | 0.6002 (± 0.0013) | 0.6113 (± 0.0054) | 0.6602 (± 0.0009) | 0.5997 (± 0.0097) | 0.4662 (± 0.0092) |

**Acknowledgments:** I would like to take this opportunity to express my profound gratitude and deep regard to Youcef Djenouri and Sohag Kabir, for their help throughout the duration of the project.

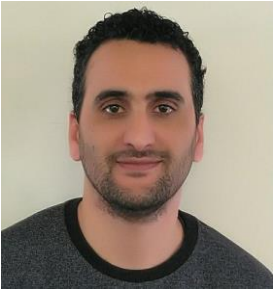

**Youcef Gheraibia:** Received the Ph.D. degree in Computer Science from the University of Annaba in 2016, and the M.Sc. degree in embedded systems from the University of OEB in 2012. He is currently; a research associate at the Dependable Intelligent Systems (DEIS) Research Group, University of Hull. He has worked as a lecturer at the University of Souk Ahras, Algeria. His research interests include model-based safety assessment, probabilistic risk and safety analysis, Machine learning, optimisation, and stochastic modelling and analysis.

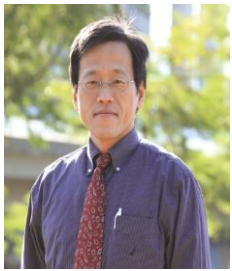

**Peng-Yeng Yin** received his B.S., M.S. and Ph.D. degrees in Computer Science from National Chiao Tung University, Hsinchu, Taiwan. From 1993 to 1994, he was a visiting scholar at the Department of Electrical Engineering, University of Maryland, College Park, and the Department of Radiology, Georgetown University, Washington D.C. In 2000, he was a visiting Professor in the Visualization and Intelligent Systems Laboratory (VISLab) at the Department of Electrical Engineering, University of California, Riverside (UCR). From 2006 to 2007, he was a visiting Professor at Leeds School of Business, University of Colorado. And in 2015, he was a visiting Professor at Graduate School of Engineering, Osaka University, Japan. He is currently a Distinguished Professor of the Department of Information Management, National Chi Nan University, Taiwan, and he was the department head during 2004 and 2006, and the Dean of the office of R&D for the university from 2008 to 2012. Dr. Yin received the Overseas Research Fellowship from Ministry of Education in 1993, Overseas Research Fellowships from National Science Council in 2000 and 2015. He is a member of the Phi Tau Phi Scholastic Honor Society and listed in Who's Who in the World, Who's Who in Science and Engineering, and Who's Who in Asia. Dr. Yin has published more than 140 academic articles in reputable journals and conferences including European Journal of Operational Research, Decision Support Systems, Annals of Operations Research, IEEE Trans. on Pattern Analysis and Machine Intelligence, IEEE Trans. on Knowledge and Data Engineering, IEEE Trans. on Education, etc. He is the Editor-in-Chief of the International Journal of Applied Metaheuristic Computing and has been on the Editorial Board of Journal of Computer Information Systems, Applied Mathematics & Information Sciences, Mathematical Problems in Engineering, International Journal of Advanced Robotic Systems, and served as a program committee member in many international conferences. He has also edited four books in pattern recognition and metaheuristic computing. His current research interests include artificial intelligence, evolutionary computation, educational informatics, metaheuristics, pattern recognition, image processing, machine learning, software engineering, computational intelligence, and operations research.

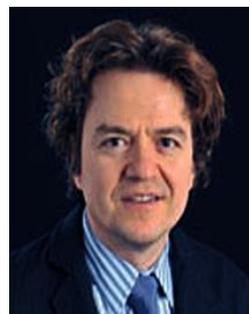

**Yiannis Papadopoulos** Professor Yiannis Papadopoulos is leader of the Dependable Systems research group at the University of Hull. He pioneered the HiP-HOPS model-based dependability analysis and optimisation method and contributed to the EAST-ADL automotive design language, working with Volvo, Honda, Continental, Honeywell and DNV-GL, among others. He is actively involved in two technical committees of IFAC (TC 1.3 & 5.1). Contact him at Y.I.Papadopoulos@hull.ac.uk.

**Abdeoulahab Moussaoui:** is Professor at Ferhat Abbas University. He received his BSc in Computer Science in 1990 from the Department of Computer Science from the University of Science and Technology of Houari Boumedienne (USTHB),

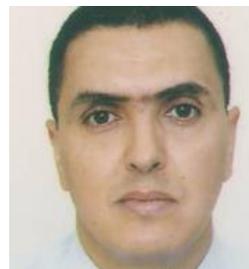

Algeria. He also received and MSc in Space Engineering in 1991 from University of Science and Technology of Oran (USTO). He received also an MSc degree in Machine Learning from Reims University (France) since 1992 and Master's degree in Computer Science in 1995 from University of Sidi Bel-abbes, Algeria and PhD degree in Computer Science from Ferhat Abbas University, Algeria where he obtains a status of full-professor in Computer Science. He is IEEE Member and AJIT, IJMMIA & IJSC

Referee. His researches are in the areas of clustering algorithms and multivariate image classification applications. His current research interests include the fuzzy neuronal network and non-parametric classification using unsupervised knowledge system applied to biomedical image segmentation and bioinformatics.

**Smaine Mazouzi**

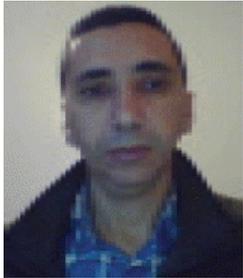

He is a professor at 20 Août 1955 University of Skikda. He received his M.S. and Ph.D. degrees in Computer Science from University of Constantine, respectively, in 1996 and 2008. His fields of interest are pattern recognition, machine vision, and computer security. His current research concerns using distributed and complex systems modeled as multi-agent systems in image understanding and intrusion detection.